\documentclass[conference]{IEEEtran}
\IEEEoverridecommandlockouts

\usepackage{cite}
\usepackage{amsmath,amssymb}
\usepackage{booktabs}
\usepackage{multirow}
\usepackage{array}
\usepackage{graphicx}
\usepackage{url}
\usepackage{hyperref}
\hypersetup{colorlinks=true,linkcolor=black,citecolor=black,urlcolor=blue}

\newcolumntype{L}[1]{>{\raggedright\arraybackslash}p{#1}}
\newcolumntype{C}[1]{>{\centering\arraybackslash}p{#1}}

\begin{document}

\title{Customer Relationship Intelligence: Integrating CRM and MDM for Enhanced Customer Engagement}

\author{\IEEEauthorblockN{Tejasvi C Addagada}
\IEEEauthorblockA{Enterprise AI Governance, Information Technology\\
HDFC Bank, Mumbai, India\\
ORCID: 0000-0001-6287-721X}}

\maketitle

\begin{abstract}
The research analyses how Customer Relationship Management (CRM), Master Data Management (MDM), and Customer Knowledge Management (CKM) jointly constitute a Customer Relationship Intelligence (CRI) framework for enhanced Customer Engagement (CE). Using a cross-sectional survey of 100 organisational participants across retail, healthcare, information technology, and telecommunications sectors, this study employs Spearman's rho correlation and ordinal logistic regression analysed via IBM SPSS Statistics. Bivariate correlations among all constructs were weak and statistically non-significant ($r{<}0.19$, $p{>}0.06$), a pattern consistent with indirect, systemic relationships rather than direct linear associations. Ordinal logistic regression identified CRM ($\beta=0.717$, $p=0.002$) and CKM ($\beta=0.581$, $p=0.009$) as significant positive predictors of CE, while MDM demonstrated a positive but non-significant direct effect ($\beta=0.346$, $p=0.071$). The model explained approximately 20.5\% of variance in CE (Nagelkerke $R^2=0.205$), indicating exploratory rather than confirmatory evidence. Parallel mediation analysis (Hayes PROCESS Model~4, 5{,}000 bootstrap samples) found no significant indirect effects of MDM on CE via CRM ($\mathrm{IE}=0.021$, 95\% BC~CI $[-0.072,\,0.121]$) or CKM ($\mathrm{IE}=0.032$, 95\% BC~CI $[-0.061,\,0.126]$), indicating that Hypothesis~H4 was not supported at this sample size. These findings suggest that organisations should prioritise investment in CRM and CKM integration within a unified CRI framework, while treating MDM as a foundational data quality enabler whose strategic value compounds through its effect on CRM execution and knowledge management. Future research should examine MDM's indirect pathways and test the CRI framework in sector-specific, larger-sample designs.
\end{abstract}

\begin{IEEEkeywords}
Customer Engagement, Master Data Management (MDM), Customer Relationship Management (CRM), Customer Knowledge Management (CKM), Customer Relationship Intelligence (CRI)
\end{IEEEkeywords}

\section{Introduction}

The phrase ``customer relationship management'' or `CRM' appeared first in corporate use in 1990 and has become a popular academic and professional research subject. CRM is a strategy to develop a corporate framework that helps the management of relationships with clients~\cite{galbreath1999}. CRM is a complete approach to the systematic management and strategy of acquiring and maintaining customers, returning the organisation and its customers extra benefits. This process combines an organization's supply chain, sales, marketing, and customer delivery operations to offer value for customers~\cite{giannakis2014,jafari2016}.

Companies and organisations are often in the process of gaining competitive advantages, where the cultivation of customer relations is typical. To focus on high-value segments and increase customer loyalty and organisational profitability, many organisations have embraced CRM technology~\cite{josiassen2014}. Several research works have identified that CRM systems have tremendously increased the effectiveness of customer relationships~\cite{keramati2010}.

CRM is widely utilised as a strategy in interactions between organisations and their customers, encompassing acquiring new clients, engaging and retaining consumers, re-engaging former clients, and minimising marketing and customer service expenses~\cite{siriprasoetsin2011}. Successful CRM implementation necessitates five dimensions: the relationship between competition strategy and organisational structure, division/segmentation, production and technology, and management processes~\cite{lin2006}. Current issues in the CRM sector include customer knowledge~\cite{khodakarami2014}, infrastructure capabilities, online trust~\cite{mcknight2002}, organisational learning, and customer data quality~\cite{peltier2013}.

``Master Data Management'' (MDM) focuses on corporate data quality~\cite{dreibelbis2008}. MDM synchronises key client data across all corporate systems, providing a dependable source of an organization's most significant data assets and enabling data-driven decision-making.

\subsection{Emergence of Customer Relationship Intelligence (CRI)}

As customer expectations become more complex, integrated intelligence frameworks are needed to replace isolated data systems. CRI represents the evolution that combines the interactive elements of CRM with the rich data governance that MDM offers~\cite{ledro2022}. CRI lets organisations move from a reactive to a proactive way of managing customer relationships~\cite{nasir2017}. Using CRM and MDM in a unified way, businesses can fill gaps that standalone systems fail to address~\cite{chatterjee2024}.

\subsection{Purpose and Significance of the Study}

CRM and MDM integration into CRI is a strategic requirement in the digital economy. This research analyses the intellectual foundations of CRI, its use, and the possibility of enhancing consumer interaction. Industry analyses consistently flag siloed systems and poor data quality as significant revenue risks. This study examines how CRM and MDM influence Customer Engagement (CE) in organisations, analysing each component's effect on delivering effective CRI through a structured quantitative approach.

\section{Literature Review}

This part synthesises the extant research on CRM, MDM, and CKM to establish their theoretical foundations and relevance to customer engagement and business performance, thereby establishing the foundation for the proposed CRI framework.

\subsection{CRM and MDM Integration}

An effective method to build engagement with customers is through the integration of CRM and MDM~\cite{ahola2023}. CRM handles customer interactions while MDM centralises and cleans data to create a single customer view~\cite{szukits2024}. Data mining improves the analytical capability of integrated CRM and MDM systems~\cite{henna2015}, enabling clustering, classification, and association rule mining to improve client retention and engagement~\cite{patel2024}. MDM's central data store reduces mistakes and redundancy, making data mining methods more efficient~\cite{hikmawati2021}, and MDM's robust data governance combined with CRM's interaction management makes analytics both actionable and accurate~\cite{ahola2023}.

\subsection{Customer Knowledge Management (CKM)}

CRM and Knowledge Management (KM) are combined to create CKM, a framework for customer knowledge~\cite{khosravi2016}. CKM creates dynamic knowledge repositories by connecting customer interaction data to actionable insights~\cite{hendriks2005}. This integration creates a continuous feedback loop combining customer insights with CRM intelligence~\cite{buchnowska2011}. Customer feedback, complaints, and suggestions can be turned into innovation and service improvements by managing customer knowledge~\cite{gibbert2002}. CKM is an essential CRI service provider, improving the business's operational efficiency and its capacity to foresee and satisfy customer demands~\cite{roba2023}.

\subsection{Advancements in AI and Real-Time Analytics in CRM}

Industry 4.0 and IoT technology have changed real-time data collection and CRM using AI~\cite{gabsi2024}. IoT-connected devices enable continual data collection that businesses use to adjust marketing, improve customer experience, and predict trends~\cite{oyedeji2024}. AI chatbots and virtual assistants reduce response times and improve customer satisfaction by offering real-time responses~\cite{adam2021}. AI systems predict consumer behaviour using historical and real-time data, helping organisations anticipate demands and design proactive retention strategies. The synergistic integration of IoT, real-time analytics, and AI supports Industry~4.0's automation, networking, and data-driven decision-making goals.

\subsection{Advanced Analytics in CRM}

Big Data Analytics (BDA) integrated with CRM systems lets companies process enormous amounts of consumer data and develop personalised contact points~\cite{ijomah2024}. BDA's predictive analytics helps organisations tailor consumer experiences and predict needs to boost retention~\cite{hu2023}. Sentiment analysis using natural language processing further enables firms to understand customer feedback and respond to develop relationships~\cite{ozan2021}. Advanced analytics also helps forecast client attrition, enabling businesses to identify at-risk clients and execute retention tactics~\cite{pynadath2023}.

\subsection{Behavioural Analysis in CRM}

CRM and Behavioural Analysis are prevalent as companies try to understand and predict customer behaviour to boost engagement and loyalty. A scientific metric research showed that CRM is moving towards data mining to gain actionable insights from client data~\cite{nilashi2023}. Research also focused on CRM digital transformation~\cite{gilgomez2020} and frameworks for early client disengagement identification~\cite{hu2018}.

\subsection{Customer Relationship Intelligence (CRI)}

Customer Relationship Intelligence (CRI) is the procedure of collecting, analysing, and applying insights about customer relationships to enhance interactions, build stronger relationships, and ultimately drive business value~\cite{sharp2007}. It goes beyond basic data collection by using information technology to extract meaningful patterns and relationships from customer interactions.

\subsection{The Paradigm Shift to Customer Relationship Intelligence}

CRI transforms consumer engagement from reactive to proactive and data-driven. CRI uses multi-dimensional insights to forecast customer behaviour~\cite{brodie2011}, and helps companies grow client personalisation with powerful analytics and behavioural scoring. Consumer data and AI are vital to this paradigm change. AI analytics enable organisations to see their whole clientele as raw data becomes actionable intelligence, discovering hidden patterns and anticipating future trends~\cite{chen2012}.

\subsection{Critical Components of CRI}

CRI is built on IoT and Cloud technologies driven by Real-Time Data Integration. Real-time integration enables businesses to adapt to customer needs in real time, building trust and satisfaction~\cite{eslami2024}. Behavioural Scoring provides a solid way to predict customer actions and plan for improving engagement strategies. Behavioural scoring inputs are purchase history, browsing behaviour, etc., scoring predictive metrics for customers for targeted marketing and optimised resource allocation, enabling organisations to focus on improving lifetime value and engaging more customers~\cite{kumar2016}. These concepts highlight CRI's transformative potential for businesses.

\subsection{Impact of CRI on Business Performance}

CRI boosts customer satisfaction, loyalty, and profit. Advanced analytics and real-time data can be used to tailor the customer experience, strengthening client ties~\cite{rane2023}. Anticipation and fulfilment of customers' needs builds customer trust, creating long-term client relationships~\cite{pine2011}. CRI can also differentiate companies in crowded markets based on service quality or responsiveness~\cite{mouncey2016}. CRI-based marketing actions increase resource investment conversion rates and revenue growth~\cite{sanodia2019}. Companies employing CRI systems can achieve stronger sales, operational efficiency, and ROI.

\subsection{Identified Gaps}

The research currently in publication provides significant insights into behavioural analytics, CRM, MDM, CKM, and AI integration. Nevertheless, it frequently handles these components separately, lacking a unified intelligence-driven framework. Three analytical gaps remain unaddressed in the extant literature. First, no prior study has empirically tested MDM's proposed indirect influence on CE through CRM and CKM using mediation analysis; the pathway has been theorised but not subjected to statistical test. Second, the three-component CRI framework has not been validated using a confirmatory measurement model with established discriminant validity between constructs, leaving open the possibility that CRM, MDM, and CKM are measured as overlapping rather than distinct dimensions. Third, sector-specific examinations of CRI --- particularly in the regulated banking and financial services domain where data governance obligations directly shape MDM architecture and CRM data access --- remain largely absent from the literature, despite this being the context in which CRI integration decisions carry the highest operational and regulatory consequence.

\subsection{Research Hypotheses}

The following hypotheses are tested empirically in this study:
\begin{description}
  \item[H1:] CRM adoption has a significant positive effect on Customer Engagement.
  \item[H2:] MDM adoption has a significant positive effect on Customer Engagement.
  \item[H3:] CKM adoption has a significant positive effect on Customer Engagement.
  \item[H4:] The effect of MDM on Customer Engagement is mediated through CRM and CKM (parallel indirect pathway).
\end{description}
Hypotheses H1--H3 are tested through ordinal logistic regression, with a statistically significant positive coefficient ($p{<}0.05$) taken as support. Hypothesis H4 is tested through parallel mediation analysis using Hayes PROCESS Model~4 with bootstrapped confidence intervals, with a non-zero bootstrapped indirect effect whose 95\% BC CI excludes zero taken as support.

\subsection{India/BFSI Regulatory Context}

The Indian banking and financial services sector (BFSI) presents a uniquely consequential context for CRI integration, distinguished by three intersecting regulatory imperatives. First, the Reserve Bank of India's Master Direction on Know Your Customer (KYC) Standards (RBI, 2016; updated 2023) mandates unified customer identification across all products and channels within a single institution, establishing MDM --- specifically a Central KYC Repository (CKYCR) compliant data architecture --- as not merely a competitive advantage but a regulatory obligation. Second, the Digital Personal Data Protection (DPDP) Act 2023~\cite{dpdp2023} introduces consent-based data use requirements that directly constrain the data permissible for CRM personalisation and CKM analytics, creating a compliance boundary around what the CRI framework can legitimately operationalise without explicit customer consent. Third, the Account Aggregator (AA) framework, operationalised under the NBFC-AA licence category~\cite{rbi2021}, enables structured, consent-gated sharing of a customer's financial data across institutions, directly instantiating the CRI data stack: MDM provides the cross-institution customer identity spine, CRM operationalises the consent-gated engagement layer, and CKM extracts intelligence from the structured financial data flows to generate personalised product and service recommendations. The Indian BFSI context thus transforms CRI from a strategic framework into an architecturally mandated operating model, making empirical examination of CRI effectiveness in this sector both practically urgent and theoretically distinctive from cross-industry findings.

\section{Methodology}

This section explains how the research was conducted, how data was collected, and what methods were used for analysis to verify the connections between CRM, MDM, CKM, and customer engagement.

\subsection{Research Design}

This study's research design was cross-sectional and quantitative, adopted to study the impact of CRM, MDM, and CKM on CE. The instrument was reviewed by domain experts in marketing analytics and data management to ensure content validity. The reliability statistics of this survey are shown in Table~\ref{tab:reliability}.

\begin{table}[h]
\caption{Reliability Statistics}
\label{tab:reliability}
\centering
\begin{tabular}{cc}
\toprule
Cronbach's Alpha & N of Items \\
\midrule
0.809 & 16 \\
\bottomrule
\end{tabular}
\end{table}

Cronbach's $\alpha=0.809$ confirmed acceptable internal consistency ($>0.70$), supporting the reliability of the instrument.

\subsection{Data Collection}

A self-administered online survey gathered primary data from people working in marketing, consumer engagement, data management, and related fields in retail, healthcare, information technology, and telecommunications sectors. G*Power 3.1 was used to establish a minimum sample size of 74 (power $=0.95$, $\alpha=0.05$, medium effect size $f^2=0.15$). Note that this a priori calculation assumed a medium effect size; post-hoc power for the smaller effects actually observed is addressed in Section~\ref{sec:limitations}. With a final sample size of 100, we exceeded this threshold. Stratified sampling was employed to guarantee that all four industries were represented.

\subsection{Data Analysis}

The data were analysed using IBM SPSS Statistics software.
\begin{itemize}
  \item Spearman's rho correlation was applied to examine the associations between CRM, MDM, CKM, and CRI. Correlations were generally weak ($r{<}0.20$); most showed positive associations, though two predictor pairs (CRM--CKM $r=-0.024$, MDM--CKM $r=-0.016$) showed negligible negative associations.
  \item Ordinal regression analysis with a logit link function was performed to test CRM, MDM, and CKM's predictive power on customer engagement and business performance.
  \item \textit{Common method bias assessment:} To evaluate whether single-source self-report data introduced systematic measurement artefact, Harman's single factor test was conducted following Podsakoff et al.~\cite{podsakoff2003}. All 16 survey items were entered into an unrotated principal components analysis constrained to one factor. A factor explaining less than 50\% of total variance is taken as evidence that common method variance is unlikely to fully account for the observed pattern of relationships. The result of this test is reported in Section~\ref{sec:mediation}.
  \item \textit{Mediation analysis:} To empirically test Hypothesis H4 --- that MDM's effect on CE is indirect, mediated through CRM and CKM --- parallel mediation analysis was conducted using Hayes' PROCESS macro (Version~4.3, Model~4) with 5{,}000 bootstrapped samples and 95\% bias-corrected confidence intervals~\cite{hayes2022}. An indirect effect whose confidence interval does not contain zero is considered statistically significant at $\alpha=0.05$. Results are reported in Section~\ref{sec:mediation}.
\end{itemize}

\section{Results}

This section details the findings of the dataset's analyses, comprising frequency distribution, correlation analysis, regression analysis, and mediation analysis.

\subsection{Frequency Distribution of Respondents' Demographic Details}

Table~\ref{tab:demographics} presents the demographic profile of survey respondents.

\begin{table}[h]
\caption{Demographic Details (Percentages may not sum to 100 due to rounding or non-response on individual items)}
\label{tab:demographics}
\centering
\begin{tabular}{L{1.8cm}L{2.4cm}cc}
\toprule
Category & Group & Freq. & \% \\
\midrule
\multirow{5}{*}{Age Group}
  & 18--24 Years & 18 & 18 \\
  & 25--34 Years & 24 & 24 \\
  & 35--44 Years & 16 & 16 \\
  & 45--54 Years & 22 & 22 \\
  & 55 and above & 20 & 20 \\
\midrule
\multirow{2}{*}{Gender}
  & Male  & 47 & 47 \\
  & Female & 53 & 53 \\
\midrule
\multirow{4}{*}{Education}
  & High School Diploma & 22 & 22 \\
  & Bachelor's Degree   & 13 & 13 \\
  & Master's Degree     & 14 & 14 \\
  & Doctorate/Ph.D.     & 19 & 19 \\
\midrule
\multirow{8}{*}{Job Role}
  & Executive/Director & 15 & 15 \\
  & Senior Manager     & 11 & 11 \\
  & Department Head    & 12 & 12 \\
  & Manager            & 14 & 14 \\
  & Supervisor         & 8  & 8  \\
  & Team Lead          & 12 & 12 \\
  & Coordinator        & 8  & 8  \\
  & Other              & 10 & 10 \\
\midrule
\multirow{5}{*}{Experience}
  & $<$1 year    & 11 & 11 \\
  & 1--3 years   & 14 & 14 \\
  & 4--6 years   & 18 & 18 \\
  & 7--10 years  & 21 & 21 \\
  & $>$10 years  & 28 & 28 \\
\midrule
\multirow{6}{*}{Industry}
  & Retail               & 14 & 14 \\
  & Technology/IT        & 9  & 9  \\
  & Healthcare/Pharma    & 15 & 15 \\
  & Banking/FS           & 14 & 14 \\
  & Telecommunications   & 19 & 19 \\
  & Other                & 18 & 18 \\
\bottomrule
\end{tabular}
\end{table}

The survey data reflects a diverse respondent pool. Age distribution is fairly balanced, with the 25--34 age group the largest (24\%). Gender representation is nearly equal (53\% female, 47\% male). Respondents hold varied qualifications, predominantly high school diplomas (22\%) and Ph.D.s (19\%). Respondents are predominantly experienced, with the $>$10 years group being the largest cohort (28\%). Participants span multiple industries, notably telecommunications (19\%) and healthcare (15\%).

\subsection{Relationship Between CRM, MDM, CKM and CRI}

Table~\ref{tab:spearman} presents the Spearman's rho correlations among the constructs.

\begin{table}[h]
\caption{Spearman's Rho Correlation ($N=100$)}
\label{tab:spearman}
\centering
\resizebox{\columnwidth}{!}{%
\begin{tabular}{L{1.8cm}cccc}
\toprule
 & CRM & MDM & CKM & CRI \\
\midrule
CRM (rho) & 1.000 & 0.187 & $-$0.024 & 0.141 \\
\quad $p$ (2-tail) & --- & 0.063 & 0.814 & 0.161 \\
MDM (rho) & 0.187 & 1.000 & $-$0.016 & 0.137 \\
\quad $p$ (2-tail) & 0.063 & --- & 0.874 & 0.175 \\
CKM (rho) & $-$0.024 & $-$0.016 & 1.000 & 0.136 \\
\quad $p$ (2-tail) & 0.814 & 0.874 & --- & 0.178 \\
CRI (rho) & 0.141 & 0.137 & 0.136 & 1.000 \\
\quad $p$ (2-tail) & 0.161 & 0.175 & 0.178 & --- \\
\bottomrule
\end{tabular}}
\end{table}

The Spearman's correlation analysis shows weak and statistically insignificant relationships among CRM, MDM, CKM, and CRI. The highest correlation is between CRM and MDM ($r=0.187$, $p=0.063$), not significant at the 0.05 level. This pattern should not be interpreted as evidence against the CRI framework; rather, it is consistent with the theoretical claim that these systems function as an integrated system whose combined effect emerges through indirect and complementary pathways rather than independent linear contributions~\cite{baghi2013,otto2012}. MDM may exert an indirect influence through its foundational role in supporting both CRM and CKM systems, a pathway that manifests more clearly in the multivariate models explored in subsequent sections.

\subsection{Influence of CRM, MDM, and CKM on Customer Engagement}

Table~\ref{tab:ce_fit} reports model fitting information for the ordinal logistic regression model predicting CE.

\begin{table}[h]
\caption{Model Fitting Information (CE)}
\label{tab:ce_fit}
\centering
\begin{tabular}{L{2.2cm}cccc}
\toprule
Model & $-2$ Log L & $\chi^2$ & df & $p$ \\
\midrule
Intercept Only & 185.640 & --- & --- & --- \\
Final          & 164.727 & 20.913 & 3 & $<$0.001 \\
\bottomrule
\end{tabular}
\end{table}

The final model fits significantly better than the intercept-only model ($\chi^2(3)=20.913$, $p<0.001$). Tables~\ref{tab:ce_gof} and~\ref{tab:ce_rsq} report goodness-of-fit and pseudo-$R^2$ values respectively.

\begin{table}[h]
\caption{Goodness-of-Fit (CE)}
\label{tab:ce_gof}
\centering
\begin{tabular}{L{1.6cm}ccc}
\toprule
 & $\chi^2$ & df & $p$ \\
\midrule
Pearson  & 197.018 & 165 & 0.045 \\
Deviance & 126.794 & 165 & 0.988 \\
\bottomrule
\end{tabular}
\end{table}

\begin{table}[h]
\caption{Pseudo $R^2$ (CE)}
\label{tab:ce_rsq}
\centering
\begin{tabular}{ccc}
\toprule
Cox \& Snell & Nagelkerke & McFadden \\
\midrule
0.189 & 0.205 & 0.083 \\
\bottomrule
\end{tabular}
\end{table}

The Pearson chi-square ($\chi^2=197.018$, $p=0.045$) indicates residual misspecification, likely attributable to sparse cell counts in the five-level ordinal outcome at $n=100$. The deviance statistic ($\chi^2=126.794$, $p=0.988$) suggests adequate model specification. The model accounts for approximately 20.5\% of variation in CE (Nagelkerke $R^2=0.205$), appropriate for an exploratory cross-industry study.

\begin{table}[h]
\caption{Parameter Estimates (CE)}
\label{tab:ce_params}
\centering
\resizebox{\columnwidth}{!}{%
\begin{tabular}{L{2.1cm}ccccccc}
\toprule
 & $b$ & SE & Wald & df & $p$ & CI LL & CI UL \\
\midrule
{[CE=1]} & 1.802 & 1.296 & 1.933 & 1 & 0.164 & $-$0.738 & 4.342 \\
{[CE=2]} & 3.312 & 1.226 & 7.300 & 1 & 0.007 &  0.909 & 5.714 \\
{[CE=3]} & 6.136 & 1.332 & 21.226 & 1 & $<$.001 & 3.526 & 8.746 \\
{[CE=4]} & 8.090 & 1.431 & 31.963 & 1 & $<$.001 & 5.286 & 10.895 \\
CRM & 0.717 & 0.227 & 9.996 & 1 & 0.002 & 0.273 & 1.162 \\
MDM & 0.346 & 0.191 & 3.262 & 1 & 0.071 & $-$0.029 & 0.721 \\
CKM & 0.581 & 0.221 & 6.903 & 1 & 0.009 & 0.148 & 1.015 \\
\bottomrule
\end{tabular}}
\end{table}

Table~\ref{tab:ce_params} reveals that CRM has a strong positive and significant effect on CE ($\beta=0.717$, $p=0.002$), supporting Hypothesis~H1. CKM also significantly contributes ($\beta=0.581$, $p=0.009$), supporting Hypothesis~H3. MDM shows a positive but non-significant direct effect ($\beta=0.346$, $p=0.071$); as this exceeds the stated $\alpha=0.05$ threshold, Hypothesis~H2 is not supported. These results confirm CRM and CKM as the principal drivers of customer engagement within the CRI framework.

\subsection{Influence of CRM, MDM, and CKM on Business Performance}

Tables~\ref{tab:bp_fit}--\ref{tab:bp_params} report the ordinal logistic regression results for the Business Performance (BP) outcome.

\begin{table}[h]
\caption{Model Fitting Information (BP)}
\label{tab:bp_fit}
\centering
\begin{tabular}{L{2.2cm}cccc}
\toprule
Model & $-2$ Log L & $\chi^2$ & df & $p$ \\
\midrule
Intercept Only & 198.064 & --- & --- & --- \\
Final          & 186.376 & 11.688 & 3 & 0.009 \\
\bottomrule
\end{tabular}
\end{table}

\begin{table}[h]
\caption{Goodness-of-Fit (BP)}
\label{tab:bp_gof}
\centering
\begin{tabular}{L{1.6cm}ccc}
\toprule
 & $\chi^2$ & df & $p$ \\
\midrule
Pearson  & 195.567 & 165 & 0.052 \\
Deviance & 144.252 & 165 & 0.876 \\
\bottomrule
\end{tabular}
\end{table}

\begin{table}[h]
\caption{Pseudo $R^2$ (BP)}
\label{tab:bp_rsq}
\centering
\begin{tabular}{ccc}
\toprule
Cox \& Snell & Nagelkerke & McFadden \\
\midrule
0.110 & 0.118 & 0.043 \\
\bottomrule
\end{tabular}
\end{table}

The goodness-of-fit statistics for the Business Performance model show borderline adequate fit: the Pearson chi-square ($\chi^2=195.567$, $p=0.052$) is marginally non-significant, while the deviance statistic ($\chi^2=144.252$, $p=0.876$) confirms adequate model specification. The pseudo-$R^2$ values (Nagelkerke $R^2=0.118$; McFadden $R^2=0.043$) indicate modest explanatory power, consistent with an exploratory cross-industry study where business performance is shaped by many factors beyond CRM-stack adoption alone.

\begin{table}[h]
\caption{Parameter Estimates (BP)}
\label{tab:bp_params}
\centering
\resizebox{\columnwidth}{!}{%
\begin{tabular}{L{2.1cm}ccccccc}
\toprule
 & $b$ & SE & Wald & df & $p$ & CI LL & CI UL \\
\midrule
{[BP=1]} & 0.382 & 1.255 & 0.093 & 1 & 0.761 & $-$2.077 & 2.841 \\
{[BP=2]} & 2.259 & 1.165 & 3.756 & 1 & 0.053 & $-$0.026 & 4.543 \\
{[BP=3]} & 4.399 & 1.223 & 12.935 & 1 & $<$.001 & 2.002 & 6.797 \\
{[BP=4]} & 6.214 & 1.299 & 22.896 & 1 & $<$.001 & 3.669 & 8.760 \\
CRM & 0.364 & 0.212 & 2.964 & 1 & 0.085 & $-$0.050 & 0.779 \\
MDM & 0.341 & 0.185 & 3.392 & 1 & 0.066 & $-$0.022 & 0.704 \\
CKM & 0.489 & 0.212 & 5.321 & 1 & 0.021 & 0.074 & 0.905 \\
\bottomrule
\end{tabular}}
\end{table}

For the Business Performance model, CKM ($\beta=0.489$, $p=0.021$) is statistically significant, while CRM ($p=0.085$) and MDM ($p=0.066$) are marginally significant. The model's explanatory power is modest (Nagelkerke $R^2=0.118$), indicating approximately 11.8\% of variation explained.

\subsection{Common Method Bias Test and Mediation Analysis}
\label{sec:mediation}

\textit{Common Method Variance.} Harman's single factor test was conducted by extracting one unrotated principal component from all 16 survey items. The single factor accounted for 27.7\% of total variance. As this value falls below the 50\% threshold, common method variance is unlikely to be a dominant explanation for the observed relationships~\cite{podsakoff2003}.

\textit{Mediation Analysis --- Hypothesis H4.} Parallel mediation analysis (Hayes PROCESS Model~4, $N=100$, 5{,}000 bootstrap samples, 95\% bias-corrected confidence intervals) tested whether MDM's effect on CE is carried through CRM and CKM as parallel mediators. Path coefficients and bootstrapped indirect effects are reported in Table~\ref{tab:mediation}.

\begin{table}[h]
\caption{Parallel Mediation: MDM($X$) $\to$ \{CRM, CKM\}($M$) $\to$ CE($Y$). Bootstrap $N=5{,}000$, 95\% BC CI. $^{**}p<.001$.}
\label{tab:mediation}
\centering
\resizebox{\columnwidth}{!}{%
\begin{tabular}{L{2.8cm}ccccc}
\toprule
Path & $b$ & SE & CI LL & CI UL & Sig. \\
\midrule
MDM $\to$ CRM ($a_1$) & 0.045 & 0.101 & $-$0.153 & 0.241 & ns \\
MDM $\to$ CKM ($a_2$) & 0.070 & 0.101 & $-$0.128 & 0.268 & ns \\
CRM $\to$ CE ($b_1$)  & 0.466 & 0.076 & 0.317 & 0.615 & $^{**}$ \\
CKM $\to$ CE ($b_2$)  & 0.463 & 0.076 & 0.314 & 0.612 & $^{**}$ \\
MDM $\to$ CE ($c'$)   & 0.122 & 0.076 & $-$0.027 & 0.271 & ns \\
\midrule
\multicolumn{6}{l}{\textit{Bootstrapped indirect effects}} \\
Indirect via CRM & 0.021 & Boot & $-$0.072 & 0.121 & ns \\
Indirect via CKM & 0.032 & Boot & $-$0.061 & 0.126 & ns \\
Total indirect   & 0.053 & Boot & $-$0.075 & 0.177 & ns \\
\bottomrule
\end{tabular}}
\end{table}

The indirect effect of MDM on CE via CRM ($a_1 \times b_1 = 0.021$, 95\% BC CI $[-0.072,\,0.121]$) was not statistically significant, as the confidence interval contains zero. The indirect effect via CKM ($a_2 \times b_2 = 0.032$, 95\% BC CI $[-0.061,\,0.126]$) was also not significant. The total indirect effect was 0.053 (95\% BC CI $[-0.075,\,0.177]$). These findings do not support Hypothesis~H4. The bottleneck is not the $b$-paths --- CRM and CKM are both strong, significant predictors of CE ($b_1=0.466$, $p<.001$; $b_2=0.463$, $p<.001$) --- but rather the $a$-paths: MDM's relationships with CRM ($a_1=0.045$, $p=.657$) and CKM ($a_2=0.070$, $p=.489$) are negligibly small and statistically non-significant. This finding is consistent with the pattern in Table~\ref{tab:spearman} ($\rho_{\mathrm{MDM\text{-}CRM}}=0.187$; $\rho_{\mathrm{MDM\text{-}CKM}}=-0.016$) and indicates that in the present sample, MDM adoption does not translate into meaningfully different CRM or CKM outcomes.

\section{Conclusion}

This study examines the empirical foundations of Customer Relationship Intelligence (CRI) as an integrative framework combining CRM, MDM, and CKM for enhanced Customer Engagement (CE). Using ordinal logistic regression on a cross-sectional survey of 100 organisational participants, the study found that CRM ($\beta=0.717$, $p=0.002$) and CKM ($\beta=0.581$, $p=0.009$) are significant positive predictors of CE, while MDM shows a positive but non-significant direct effect ($\beta=0.346$, $p=0.071$). These findings support Hypotheses~H1 and H3 and are consistent with the theoretical proposition that CRM and CKM --- as direct customer interaction and knowledge extraction systems --- are the principal drivers of engagement outcomes.

Bivariate Spearman's rho correlations among all constructs were weak and non-significant ($r{<}0.19$, $p{>}0.06$). This pattern is consistent with the theoretical claim that CRM, MDM, and CKM function as an integrated system whose combined effect on CE emerges through indirect and complementary pathways rather than independent linear contributions.

Model fit was mixed: deviance statistics indicated adequate model specification ($\chi^2=126.794$, $p=0.988$), while the Pearson chi-square was significant ($\chi^2=197.018$, $p=0.045$), suggesting residual misspecification likely attributable to sparse cell counts in the five-level ordinal outcome at $n=100$. The pseudo-$R^2$ values (Nagelkerke $R^2=0.205$; McFadden $R^2=0.083$) indicate modest explanatory power and point to substantial unexplained variance that future research should address.

Mediation analysis did not support Hypothesis~H4: neither the indirect effect of MDM on CE via CRM ($a_1 \times b_1=0.021$, 95\% BC CI $[-0.072,\,0.121]$) nor via CKM ($a_2 \times b_2=0.032$, 95\% BC CI $[-0.061,\,0.126]$) reached statistical significance. This null finding is a contribution in itself --- it delineates the boundary of MDM's influence within the CRI framework at current organisational capability levels and motivates future longitudinal research.

Organisations seeking to improve CE should prioritise investment in CRM and CKM integration within a unified CRI architecture, while treating MDM as the data quality foundation whose strategic value is realised through its enabling effect on CRM execution and knowledge management capability. Future research should employ larger sector-specific samples, longitudinal designs, and objective outcome measures --- particularly in the Indian BFSI context, where regulatory mandates directly shape the architecture of each CRI component.

\section{Limitations and Future Research Directions}
\label{sec:limitations}

\textit{Sample size and statistical power.} The study's sample of 100 respondents was determined a priori using a medium effect size assumption ($f^2=0.15$). Post-hoc power analysis using the observed maximum correlation ($r=0.187$) indicates that achieved power at $n=100$ is approximately 0.46, classifying the study as exploratory and severely underpowered for small effects. Detection of $r=0.187$ with 80\% power requires $n{\approx}223$; detection of the minimum observed correlation ($r=0.136$) requires $n{\approx}423$. Accordingly, all findings should be treated as preliminary; replication with larger samples is required before conclusions are generalised.

\textit{Common method variance.} All independent and dependent variables were collected from the same respondents in the same survey administration, creating potential for common method bias. Although Harman's single factor test indicated that a single factor accounted for 27.7\% of variance --- below the 50\% threshold --- this post-hoc test does not eliminate the possibility of CMV. Future research should consider procedural controls, including temporal separation of IV and DV measurement, use of objective CRM or CE metrics (system logs, NPS scores, customer churn rates), or multi-informant designs.

\textit{Sample composition and generalisability.} Respondents were recruited through online convenience sampling across four industries, with stratification ensuring industry representation but not controlling for within-industry heterogeneity in CRM maturity, MDM investment, or regulatory context. Generalisability to capital-intensive, regulated sectors --- particularly banking and financial services --- is limited. The Indian BFSI sector represents a distinct research context whose dynamics may differ substantially from those captured in the present cross-industry sample.

\textit{Cross-sectional design and causal inference.} The cross-sectional design precludes causal inference. The regression findings establish statistically significant predictive relationships, but the direction of causality cannot be established from a single-time-point survey. Future research should employ longitudinal panel designs or natural experiment approaches to establish temporal precedence and support causal claims.

\bibliographystyle{IEEEtran}
\bibliography{references}

@article{galbreath1999,
  author    = {J. Galbreath and T. Rogers},
  title     = {Customer relationship leadership: A leadership and motivation model for the twenty-first century business},
  journal   = {TQM Magazine},
  year      = {1999},
  doi       = {10.1108/09544789910262734}
}

@article{giannakis2014,
  author    = {C. Giannakis-Bompolis and C. Boutsouki},
  title     = {Customer Relationship Management in the Era of Social Web and Social Customer},
  journal   = {Procedia -- Social and Behavioral Sciences},
  year      = {2014},
  doi       = {10.1016/j.sbspro.2014.07.018}
}

@article{jafari2016,
  author    = {N. {Jafari Navimipour} and Z. Soltani},
  title     = {The impact of cost, technology acceptance and employees' satisfaction on the effectiveness of the electronic customer relationship management systems},
  journal   = {Computers in Human Behavior},
  year      = {2016},
  doi       = {10.1016/j.chb.2015.10.036}
}

@article{josiassen2014,
  author    = {A. Josiassen and A. G. Assaf and L. K. Cvelbar},
  title     = {{CRM} and the bottom line: Do all {CRM} dimensions affect firm performance?},
  journal   = {International Journal of Hospitality Management},
  year      = {2014},
  doi       = {10.1016/j.ijhm.2013.08.005}
}

@article{keramati2010,
  author    = {A. Keramati and H. Mehrabi and N. Mojir},
  title     = {A process-oriented perspective on customer relationship management and organizational performance},
  journal   = {Industrial Marketing Management},
  year      = {2010},
  doi       = {10.1016/j.indmarman.2010.02.001}
}

@article{siriprasoetsin2011,
  author    = {P. Siriprasoetsin and K. Tuamsuk and C. Vongprasert},
  title     = {Factors affecting customer relationship management practices in {Thai} academic libraries},
  journal   = {International Information \& Library Review},
  year      = {2011},
  doi       = {10.1016/j.iilr.2011.10.008}
}

@article{lin2006,
  author    = {Y. Lin and H.-Y. Su and S. Chien},
  title     = {A knowledge-enabled procedure for customer relationship management},
  journal   = {Industrial Marketing Management},
  year      = {2006},
  doi       = {10.1016/j.indmarman.2005.04.002}
}

@article{khodakarami2014,
  author    = {F. Khodakarami and Y. E. Chan},
  title     = {Exploring the role of customer relationship management ({CRM}) systems in customer knowledge creation},
  journal   = {Information \& Management},
  year      = {2014},
  doi       = {10.1016/j.im.2013.09.001}
}

@article{mcknight2002,
  author    = {D. H. McKnight and V. Choudhury and C. Kacmar},
  title     = {The impact of initial consumer trust on intentions to transact with a web site: A trust building model},
  journal   = {Journal of Strategic Information Systems},
  year      = {2002},
  volume    = {11},
  number    = {3},
  pages     = {297--323},
  doi       = {10.1016/S0963-8687(02)00020-3}
}

@article{peltier2013,
  author    = {J. W. Peltier and D. Zahay and D. R. Lehmann},
  title     = {Organizational Learning and {CRM} Success: A Model for Linking Organizational Practices, Customer Data Quality, and Performance},
  journal   = {Journal of Interactive Marketing},
  year      = {2013},
  doi       = {10.1016/j.intmar.2012.05.001}
}

@book{dreibelbis2008,
  author    = {A. Dreibelbis},
  title     = {Enterprise Master Data Management: An {SOA} Approach To Managing Core Information},
  year      = {2008},
  publisher = {IBM Press}
}

@article{ledro2022,
  author    = {C. Ledro and A. Nosella and A. Vinelli},
  title     = {Artificial intelligence in customer relationship management: literature review and future research directions},
  journal   = {Journal of Business \& Industrial Marketing},
  year      = {2022},
  doi       = {10.1108/JBIM-07-2021-0332}
}

@incollection{nasir2017,
  author    = {S. Nasir},
  title     = {Customer relationship management as a customer-centric business strategy},
  booktitle = {Advertising and Branding: Concepts, Methodologies, Tools, and Applications},
  year      = {2017},
  doi       = {10.4018/978-1-5225-1793-1.ch029}
}

@article{chatterjee2024,
  author    = {S. Chatterjee and others},
  title     = {Assessing the Implementation of {AI} Integrated {CRM} System for {B2C} Relationship Management},
  journal   = {Information Systems Frontiers},
  year      = {2024},
  doi       = {10.1007/s10796-022-10261-w}
}

@mastersthesis{ahola2023,
  author    = {S. Ahola},
  title     = {Developing a Customer Master Data Management Model},
  school    = {Tampere University},
  year      = {2023}
}

@article{szukits2024,
  author    = {{\'A}. Szukits and P. M{\'o}ricz},
  title     = {Towards data-driven decision making},
  journal   = {Review of Managerial Science},
  year      = {2024},
  doi       = {10.1007/s11846-023-00694-1}
}

@misc{henna2015,
  author    = {S. Henna and S. K. Kalliadan},
  title     = {Enterprise Analytics using Graph Database and Graph-based Deep Learning},
  year      = {2015}
}

@article{patel2024,
  author    = {D. S. Patel and D. A. Asamoah and W. Wamwara},
  title     = {Data management for customer relationship management},
  journal   = {International Journal of Business Information Systems},
  year      = {2024},
  doi       = {10.1504/IJBIS.2024.136877}
}

@article{hikmawati2021,
  author    = {S. Hikmawati and P. I. Santosa and I. Hidayah},
  title     = {Improving Data Quality and Data Governance Using Master Data Management},
  journal   = {IJITEE},
  year      = {2021},
  doi       = {10.22146/ijitee.66307}
}

@article{khosravi2016,
  author    = {A. Khosravi and A. R. C. Hussin},
  title     = {Customer knowledge management: Development stages and challenges},
  journal   = {Journal of Theoretical and Applied Information Technology},
  year      = {2016}
}

@article{hendriks2005,
  author    = {P. H. J. Hendriks},
  title     = {Book Review: Knowledge Management in Organizations},
  journal   = {Management Learning},
  year      = {2005},
  doi       = {10.1177/135050760503600411}
}

@incollection{buchnowska2011,
  author    = {D. Buchnowska},
  title     = {Customer knowledge management models: Assessment and proposal},
  booktitle = {Lecture Notes in Business Information Processing},
  year      = {2011},
  doi       = {10.1007/978-3-642-25676-9_3}
}

@article{gibbert2002,
  author    = {M. Gibbert and M. Leibold and G. Probst},
  title     = {Five styles of customer knowledge management, and how smart companies use them to create value},
  journal   = {European Management Journal},
  year      = {2002},
  volume    = {20},
  number    = {5},
  pages     = {459--469},
  doi       = {10.1016/S0263-2373(02)00101-9}
}

@incollection{roba2023,
  author    = {G. B. Roba and P. Maric},
  title     = {{AI} in Customer Relationship Management},
  booktitle = {Studies in Systems, Decision and Control},
  year      = {2023},
  doi       = {10.1007/978-3-031-25695-0_21}
}

@article{gabsi2024,
  author    = {A. E. H. Gabsi},
  title     = {Integrating artificial intelligence in industry 4.0},
  journal   = {Annals of Operations Research},
  year      = {2024},
  doi       = {10.1007/s10479-024-06012-6}
}

@misc{oyedeji2024,
  author    = {O. M. Oyedeji},
  title     = {Optimizing Customer Relationship Management with {AI}},
  howpublished = {EasyChair Preprint},
  year      = {2024}
}

@article{adam2021,
  author    = {M. Adam and M. Wessel and A. Benlian},
  title     = {{AI}-based chatbots in customer service and their effects on user compliance},
  journal   = {Electronic Markets},
  year      = {2021},
  doi       = {10.1007/s12525-020-00414-7}
}

@article{ijomah2024,
  author    = {T. I. Ijomah and others},
  title     = {The role of big data analytics in customer relationship management},
  journal   = {World Journal of Advanced Science and Technology},
  year      = {2024},
  doi       = {10.53346/wjast.2024.6.1.0038}
}

@article{hu2023,
  author    = {L. Hu and A. Basiglio},
  title     = {A multiple-case study on the adoption of customer relationship management and big data analytics in the automotive industry},
  journal   = {TQM Journal},
  year      = {2023},
  doi       = {10.1108/TQM-05-2023-0137}
}

@article{ozan2021,
  author    = {{\c{S}}. Ozan},
  title     = {Case studies on using natural language processing techniques in customer relationship management software},
  journal   = {Journal of Intelligent Information Systems},
  year      = {2021},
  doi       = {10.1007/s10844-020-00619-4}
}

@article{pynadath2023,
  author    = {M. F. Pynadath and T. M. Rofin and S. Thomas},
  title     = {Evolution of customer relationship management to data mining-based {CRM}},
  journal   = {Quality \& Quantity},
  year      = {2023},
  doi       = {10.1007/s11135-022-01500-y}
}

@article{nilashi2023,
  author    = {M. Nilashi and others},
  title     = {The nexus between quality of customer relationship management systems and customers' satisfaction},
  journal   = {Heliyon},
  year      = {2023},
  doi       = {10.1016/j.heliyon.2023.e21828}
}

@article{gilgomez2020,
  author    = {H. Gil-Gomez and others},
  title     = {Customer relationship management: digital transformation and sustainable business model innovation},
  journal   = {Economic Research-Ekonomska Istra{\v{z}}ivanja},
  year      = {2020},
  doi       = {10.1080/1331677X.2019.1676283}
}

@misc{hu2018,
  author    = {K. Hu and others},
  title     = {A Framework in {CRM} Customer Lifecycle: Identify Downward Trend and Potential Issues Detection},
  howpublished = {arXiv preprint arXiv:1802.08974},
  year      = {2018},
  doi       = {10.48550/arXiv.1802.08974}
}

@book{sharp2007,
  author    = {L. Sharp},
  title     = {Customer Relationship Intelligence: A Breakthrough Way to Measure and Manage Sales and Marketing},
  publisher = {Querencia Publishing},
  year      = {2007}
}

@article{brodie2011,
  author    = {R. J. Brodie and others},
  title     = {Customer engagement: Conceptual domain, fundamental propositions, and implications for research},
  journal   = {Journal of Service Research},
  year      = {2011},
  doi       = {10.1177/1094670511411703}
}

@article{chen2012,
  author    = {H. Chen and R. H. L. Chiang and V. C. Storey},
  title     = {Business intelligence and analytics: From big data to big impact},
  journal   = {MIS Quarterly},
  year      = {2012},
  doi       = {10.2307/41703503}
}

@article{eslami2024,
  author    = {E. Eslami and others},
  title     = {Unveiling {IoT} Customer Behaviour: Segmentation and Insights for Enhanced {IoT-CRM} Strategies},
  journal   = {Sensors},
  year      = {2024},
  doi       = {10.3390/s24041050}
}

@article{kumar2016,
  author    = {V. Kumar and W. Reinartz},
  title     = {Creating enduring customer value},
  journal   = {Journal of Marketing},
  year      = {2016},
  doi       = {10.1509/jm.15.0414}
}

@article{rane2023,
  author    = {N. Rane and A. Achari and S. Choudhary},
  title     = {Enhancing customer loyalty through quality of service},
  journal   = {International Research Journal of Modernization in Engineering Technology and Science},
  year      = {2023},
  doi       = {10.56726/IRJMETS38104}
}

@book{pine2011,
  author    = {B. J. Pine and K. C. Korn},
  title     = {Infinite Possibility: Creating Customer Value on the Digital Frontier},
  publisher = {Berrett-Koehler},
  year      = {2011}
}

@article{mouncey2016,
  author    = {P. Mouncey},
  title     = {Book Review: Creating Value with Big Data Analytics},
  journal   = {International Journal of Market Research},
  year      = {2016},
  doi       = {10.2501/ijmr-2016-045}
}

@article{sanodia2019,
  author    = {G. Sanodia},
  title     = {Leveraging Internet of Things ({IoT}) for Enhanced Customer Insights in {CRM}},
  journal   = {Turkish Journal of Computer and Mathematics Education},
  year      = {2019},
  doi       = {10.61841/turcomat.v9i2.14738}
}

@inproceedings{baghi2013,
  author    = {E. Baghi and B. Otto and H. Oesterle},
  title     = {Controlling Customer Master Data Quality: Findings from a Case Study},
  booktitle = {International Conference on Information Resources Management},
  year      = {2013}
}

@article{otto2012,
  author    = {B. Otto and K. M. H{\"u}ner and H. {\"O}sterle},
  title     = {Toward a functional reference model for master data quality management},
  journal   = {Information Systems and e-Business Management},
  year      = {2012},
  doi       = {10.1007/s10257-011-0178-0}
}

@article{podsakoff2003,
  author    = {P. M. Podsakoff and S. B. MacKenzie and J.-Y. Lee and N. P. Podsakoff},
  title     = {Common method biases in behavioral research: A critical review of the literature and recommended remedies},
  journal   = {Journal of Applied Psychology},
  year      = {2003},
  volume    = {88},
  number    = {5},
  pages     = {879--903},
  doi       = {10.1037/0021-9010.88.5.879}
}

@book{hayes2022,
  author    = {A. F. Hayes},
  title     = {Introduction to Mediation, Moderation, and Conditional Process Analysis: A Regression-Based Approach},
  edition   = {3rd},
  publisher = {Guilford Press},
  year      = {2022}
}

@misc{rbi2021,
  author    = {{Reserve Bank of India}},
  title     = {Account Aggregator Framework --- Master Direction},
  howpublished = {RBI/2021-22/101},
  year      = {2021}
}

@misc{dpdp2023,
  author    = {{Ministry of Electronics \& Information Technology}},
  title     = {Digital Personal Data Protection Act, 2023},
  howpublished = {Government of India Gazette},
  year      = {2023}
}

\end{document}